\documentclass[12pt,a4paper]{article}
\usepackage[left=30mm,right=15mm,top=20mm,bottom=20mm]{geometry}
\usepackage{fontspec}
\usepackage{amsmath,amssymb,graphicx}
\usepackage{pdflscape}
\usepackage{url}
\begin{document}
\begin{center}\fontsize{14}{17}\selectfont\bfseries RANK-TURBULENCE DELTA AND INTERPRETABLE APPROACHES TO STYLOMETRIC DELTA METRICS\end{center}

\begin{center}\fontsize{14}{17}\selectfont Dmitry Pronin, Evgeny Kazartsev\end{center}

\begin{center}\fontsize{14}{17}\selectfont HSE University\end{center}

The study was implemented in the framework of the Basic Research Program at HSE University (HSE-BR-2025-031) in 2026.

\noindent\textbf{Abstract}\par
This article introduces two new measures for authorship attribution — Rank-Turbulence Delta and Jensen–Shannon Delta — which generalise Burrows’s classical Delta by applying distance functions designed for probabilistic distributions. We first set out the theoretical basis of the measures, contrasting centred and uncentred z-scoring of word-frequency vectors and re-casting the uncentred vectors as probability distributions. Building on this representation, we develop a token-level decomposition that renders every Delta distance numerically interpretable, thereby facilitating close reading and the validation of results. The effectiveness of the methods is assessed on four literary corpora in English, German, French and Russian. The English, German and French datasets are compiled from Project Gutenberg, whereas the Russian benchmark is the SOCIOLIT corpus containing 639 works by 89 authors spanning the eighteenth to the twenty-first centuries. Rank-Turbulence Delta attains attribution accuracy comparable with Cosine Delta; Jensen–Shannon Delta consistently matches or exceeds the performance of canonical Burrows’s Delta. Finally, several established attribution algorithms are re-evaluated on the extended SOCIOLIT corpus, providing a realistic estimate of their robustness under pronounced temporal and stylistic variation.


\section*{1. Introduction}

Computational stylometry is a branch of the digital humanities devoted to the attribution of texts through the identification and quantitative analysis of their stylistic characteristics. This methodology makes it possible to identify the author of an unknown or disputed work by relying on statistically significant regularities in language use. A key theoretical premise of stylometric analysis is that an individual’s linguistic behaviour is marked by stable, largely unconscious patterns, which in turn produce stylistic consistency across texts written by the same author. Authorship determination therefore finds application in a wide array of disciplines, including literary studies, philosophy, history, and forensic linguistics.

Among the various approaches developed within stylometry, the Delta metric proposed by Burrows (Burrows  2002) remains one of the most influential and widely applied. Burrows’s Delta measures stylistic distance by comparing standardized (z-scored) word-frequency profiles across texts. Since its introduction, Delta has been employed in a wide range of linguistic and literary contexts and has generated a substantial body of methodological reflection. Foundational analyses of its geometric and probabilistic properties have been provided by Argamon (Argamon 2008) and Evert (Evert et al. 2017), while numerous modifications have been proposed in order to improve robustness or adapt the metric to specific tasks, including Hoover’s Delta (Hoover 2004), Eder’s Delta (Eder, Rybicki and Kestemont 2016), Rotated Delta (Jannidis et al. 2015), and Cosine Delta. Comparative evaluations demonstrate that no single variant dominates across all corpora and evaluation protocols; performance depends on language, corpus composition, feature selection, and task design (Evert et al. 2017; Proisl et al. 2018; Kocher and Savoy 2019).

At the same time, it is important to situate Delta within the broader landscape of authorship attribution. Contemporary attribution systems are predominantly based on supervised machine-learning approaches, ranging from linear classifiers and support vector machines to neural architectures (Stamatatos 2009; Kestemont et al. 2019; Bevendorff et al. 2021; Kestemont et al. 2021; Cafiero and Camps 2023). These systems are typically optimized for predictive accuracy under carefully controlled evaluation settings. The aim of the present article is not to compete with such supervised pipelines. Rather, we focus on improving and theoretically reinterpreting the family of Delta-like distance measures that remain central to exploratory and unsupervised stylometric research.

Distance-based stylometry continues to play a crucial role in clustering, network visualization, rolling stylometry, corpus exploration, diachronic comparison, and non-attributional stylistic analysis (Eder 2016; Eder 2017; Cafiero and Camps 2019; Camps et al. 2021; Beausang 2022; Päpcke et al. 2023; Nilsson-Fernàndez 2025). In such contexts, interpretability, theoretical transparency, and stability under variation are often as important as raw classification performance. Delta occupies an intermediate methodological position: it is statistically effective, yet its operating principle remains linguistically interpretable and historically embedded in philological traditions. Recent work has emphasized this dual character and called for a more explicit linguistic interpretation of Delta (Kovalev and Dviniatin 2025; Kovalev 2024).

In much of exploratory stylometry, interpretability refers to the possibility of relating quantitative distance patterns to historically and linguistically meaningful stylistic tendencies. However, the contribution of individual tokens to a given Delta value is rarely formalized in a systematic way. In the present article, we try to adopt a more specific notion of interpretability, understood as the identification and numerical decomposition of token-level contributions to the overall distance. Such a procedure does not replace broader philological interpretation, but provides a transparent bridge between aggregate distance measures and the lexical features that generate them. It may also offer empirical guidance when selecting the number of most frequent words and assessing the stability of results under variation.

To achieve this goal, we offer a theoretical reinterpretation of Burrows’s Delta by examining both centred and uncentred standardized frequency vectors. We show that uncentred z-vectors can be normalized and interpreted as probability distributions, thereby opening the possibility of employing distance measures drawn from information theory and complex-systems analysis, such as Jensen–Shannon divergence and Rank-Turbulence divergence (Dodds et al. 2023). Second, we develop a systematic token-level decomposition for Delta and its probabilistic variants, enabling the numerical attribution of stylistic distance to individual lexical items.

The proposed probabilistic methods are evaluated on multilingual corpora in English, German, French, and Russian. Particular attention is given to the part of Russian-language SOCIOLIT corpus, which comprises 639 works by 89 authors spanning the eighteenth to the twenty-first centuries. Delta and its variants have been actively applied to Russian literary material in recent years (Orekhov 2021; Velikanova and Orekhov 2019; Dviniatin and Kovalev 2024), demonstrating both the methodological relevance of distance-based stylometry and the need for theoretically grounded interpretative frameworks. By testing our measures under substantial temporal and stylistic variation, we aim to assess their robustness in realistic exploratory settings.

In what follows, we first formalize Delta in vector-space terms and distinguish centred and uncentred standardizations. We then reinterpret uncentred representations as probability and rank distributions, introduce corresponding distance measures, and derive token-level decompositions. The experimental sections evaluate both clustering and attribution performance, as well as the interpretative potential of the proposed framework.

\section*{2. Delta as a Metric in Word-Vector Space}

Since its introduction, Burrows’s Delta has been interpreted both as a heuristic measure of stylistic distance and as a formally defined metric in a standardized frequency space (Argamon 2008; Evert et al. 2017). Making this geometric interpretation explicit clarifies the relationship between the classical formulation and its later modifications. Canonically, the Burrows’s Delta distance between two texts is given by:

\begin{equation}\tag{1}
\Delta(T_1,T_2)=\frac1n\sum_{i=1}^{n}\left|z_i^{(1)}-z_i^{(2)}\right|
\end{equation}

where $n$ is the number of tokens in the corpus selected for analysis, $z_i^{(1)},z_i^{(2)}$ are the standardized scores (z-scores) of the relative frequencies of token $\tau_i$ in texts $T_1,T_2$ respectively. These z-scores are computed as:

\begin{equation}\tag{2}
z_\tau^{(i)}=\frac{p_\tau^{(i)}-\mu_\tau}{\sigma_\tau}
\end{equation}

in which $p_\tau^{(i)}$ - denotes the relative frequency (probability) of token $\tau$ in i-th text, $\mu_\tau$ is the mean relative frequency of token $\tau$ across the entire corpus and $\sigma_\tau$ its standard deviation. Tokens may be single words as well as lexical or even character n-grams; nevertheless, in most studies the n-grams in question are words.

Viewing Burrows’s Delta within a feature-space framework is convenient, because this representation provides a theoretical rationale for various modifications of the metric. If texts are embedded in a high-dimensional feature space under a ‘bag-of-words’ model, vectorization proceeds by computing the relative frequencies of a predefined token set. Applying standardization (z-scoring) to those frequencies yields a unified representation in which each document is mapped to a vector:

\begin{equation}\tag{3}
T\longrightarrow(z_1,z_2,\ldots,z_n)
\end{equation}

In this form, the Manhattan (L$_1$) distance between the z-vectors of texts is identical to the canonical Burrows’s Delta, up to the constant $n$, which can be ignored when results are compared within a fixed word set. In geometric terms, the classical Burrows’s Delta corresponds to the Manhattan (L$_1$) distance between standardized frequency vectors. Alternative distance measures offer different stylistic lenses: Euclidean (L$_2$) Delta measures the quadratic deviation across dimensions, giving greater weight to large coordinate differences; Cosine Delta, in turn, evaluates the angular similarity between vectors, focusing on the direction of stylistic deviation rather than its absolute magnitude. Such a view of Delta establishes a universal basis for a broad range of analytical procedures, including vector distances, dimensionality reduction, and clustering.

A large family of Delta variants involves token-wise differences between z-scores in which the terms responsible for mean centring are cancelled:

\begin{equation}\tag{4}
z_\tau^{(1)}-z_\tau^{(2)}=\frac{p_\tau^{(1)}-\mu_\tau}{\sigma_\tau}-\frac{p_\tau^{(2)}-\mu_\tau}{\sigma_\tau}=\frac1{\sigma_\tau}\left(p_\tau^{(1)}-p_\tau^{(2)}\right)
\end{equation}

Because of this property, one may simplify computations by using a vector of uncentred standard scores, i.e. omitting the subtraction of corpus means:

\begin{equation}\tag{5}
T\longrightarrow\left(\frac{p_1}{\sigma_1},\frac{p_2}{\sigma_2},\ldots,\frac{p_n}{\sigma_n}\right)
\end{equation}

Representing texts through uncentred z-vectors can be interpreted as weighting the dimensions of the feature space: the smaller the standard deviation of a token’s frequency in the corpus, the more influential the difference between its frequencies in different texts becomes. Low-variance tokens therefore exert a greater impact on distance calculations, because their deviations from the mean are more informative for stylometric analysis. For distance functions that depend solely on pairwise coordinate differences, the centred and uncentred representations yield equivalent results, since the centring terms cancel out.

\section*{3. Z-vectors as a Probability and Rank Distributions}

One of the key advantages of the uncentred z-vector over its centred counterpart is that all its components are non-negative. This property makes it possible to reinterpret the vector as a probability distribution. The components of such a distribution are obtained from the uncentred z-vector by normalization:

\begin{equation}\tag{6}
T\longrightarrow(\rho_1,\rho_2,\ldots,\rho_n)
\end{equation}

where

\begin{equation}\tag{7}
\rho_\tau^{(i)}=\frac{p_\tau^{(i)}}{\sigma_\tau}\left(\sum_{j=1}^{n}\frac{p_j^{(i)}}{\sigma_j}\right)^{-1}
\end{equation}

The normalization in (6) ensures that the transformed vector sums to one across all dimensions. Each component can therefore be interpreted as the relative weight of a token within the stylometric profile of the text. Importantly, the transformation preserves the relative contrasts encoded in the standardized frequencies while embedding them in a probability simplex

Gallagher (Gallagher et al. 2021) has demonstrated that, by representing texts not as points in a feature space but rather as probability distributions, it becomes possible to apply distance measures derived from probability theory. For instance, in this space the distance between two texts can be expressed by the Jensen–Shannon divergence:

\begin{equation}\tag{8}
D^{(JS)}(P^{(2)}\Vert P^{(1)})=\pi_1D^{(KL)}(P^{(1)}\Vert M)+\pi_2D^{(KL)}(P^{(2)}\Vert M)
\end{equation}

where $M=\pi_1P^{(1)}+\pi_2P^{(2)}$ is the mixed distribution. The Jensen–Shannon divergence compares each distribution not directly to the other, but to their weighted average. The mixed distribution represents a hypothetical intermediate stylistic profile situated between the two texts. In effect, the divergence measures how much information would be lost if one text were approximated by this intermediate profile. The mixture weights $\pi_1,\pi_2$ are chosen such that $\pi_1+\pi_2=1$. The weights assigned to the mixture determine the relative influence of each text in constructing this intermediate distribution. In the absence of prior assumptions, equal weights $\pi_1=\pi_2=\frac12$ are typically used, thereby treating both texts symmetrically.

The Kullback–Leibler divergence $D^{(KL)}$ elements of the Jensen–Shannon divergence, computed as follows:

\begin{equation}\tag{9}
D^{(KL)}(P\Vert M)=\sum_{i=1}^{n}\rho_i\log_2\frac{\rho_i}{m_i}
\end{equation}

In practice, zero probabilities must be handled carefully, since the logarithmic term in the Kullback–Leibler divergence is undefined for zero values. We therefore apply additive smoothing by introducing a small uniform pseudocount prior to normalization. This ensures numerical stability without materially altering the comparative structure of the distributions.

Whereas probabilistic divergences operate directly on probability distributions, rank-based representations focus on the hierarchical ordering of tokens. In many stylometric contexts, especially when corpora are heterogeneous or limited in size, the relative prominence of words may be more stable than their precise frequencies. A shift in rank reflects a reorganisation of lexical importance rather than merely a quantitative fluctuation. An effective way of comparing rank distributions is the rank-turbulence divergence proposed by Dodds (Dodds et al 2023):

\begin{equation}\tag{10}
D_\alpha^{(RT)}(R_1\Vert R_2)=\frac1{\mathcal N_{1,2;\alpha}}\frac{\alpha+1}{\alpha}\sum_{\tau\in R_{1,2}}\left|\frac1{(r_\tau^{(1)})^\alpha}-\frac1{(r_\tau^{(2)})^\alpha}\right|^{1/(\alpha+1)}
\end{equation}

The role of the normalization coefficient  $\mathcal N_{1,2;\alpha}$ is to ensure that the divergence remains comparable across datasets with different vocabulary sizes. In addition, the normalization rescales the measure to a bounded interval between 0 and 1. This boundedness is not merely a mathematical convenience: it provides an interpretable scale on which stylistic distance can be assessed independently of corpus size or lexical richness. A divergence close to 0 indicates a high degree of similarity in rank structure, whereas values approaching 1 signal substantial reorganization. Normalization coefficient $\mathcal N_{1,2;\alpha}$ is computed as:

\begin{equation}\tag{11}
\begin{aligned}\mathcal N_{1,2;\alpha}={}&\frac{\alpha+1}{\alpha}\sum_{\tau\in R_1}\left|\frac1{(r_\tau^{(1)})^\alpha}-\frac1{(N_1+0.5N_2)^\alpha}\right|^{1/(\alpha+1)}\\&+\frac{\alpha+1}{\alpha}\sum_{\tau\in R_2}\left|\frac1{(N_2+0.5N_1)^\alpha}-\frac1{(r_\tau^{(2)})^\alpha}\right|^{1/(\alpha+1)}\end{aligned}
\end{equation}

Here $N_1,N_2$ are the numbers of distinct types in each system, $\alpha$ is a user-chosen parameter, and $r_\tau^{(1)},r_\tau^{(2)}$ denote the ranks of a token in the respective distributions. Note also that a rank distribution can be derived from either an uncentred or a centred z-vector; in the latter case, any negative components (resulting from subtracting the mean) will naturally fall at the bottom of the resulting rank ordering.

The parameter $\alpha$ controls the sensitivity of the divergence to movements in different regions of the rank spectrum. Smaller values of alpha parameter ($\alpha\to0$) amplify the contribution of lower-ranked (i.e. less frequent) tokens, making the divergence sensitive to subtle lexical shifts. Larger values ($\alpha\to\infty$) concentrate attention on high-frequency items, thus emphasising dominant stylistic patterns. This flexibility allows the researcher to adjust the analytical lens depending on whether the focus lies on broad stylistic tendencies or on more marginal, potentially idiosyncratic elements.

Taken together, the probabilistic and rank-based representations extend the geometric interpretation of Delta without abandoning its core principle: stylistic distance remains grounded in systematic contrasts between frequency profiles, but can now be examined through alternative comparison geometries.

\section*{4. Interpretation of Distances}

Although previous studies have discussed the geometric and probabilistic foundations of Delta (Argamon 2008; Evert et al. 2017), the contribution of individual tokens to the overall distance has rarely been formalized in a systematic and reproducible manner. The present section develops such a decomposition framework, enabling each distance measure to be expressed as a sum of token-level contributions.

Formally, denoting by $\delta\Delta_\tau=\left|\frac{p_\tau^{(1)}}{\sigma_\tau}-\frac{p_\tau^{(2)}}{\sigma_\tau}\right|$ the contribution of token $\tau$ to the Burrows’s Delta distance between texts, we write:

\begin{equation}\tag{12}
\Delta(T_1,T_2)=\frac1n\sum_{i=1}^{n}\delta\Delta_i
\end{equation}

Every $\delta\Delta_\tau\geq0$; hence $\delta\Delta_\tau$ measures the extent to which token $\tau$ is over-represented in one text relative to the other. Specifically, if $\frac{p_\tau^{(1)}}{\sigma_\tau}>\frac{p_\tau^{(2)}}{\sigma_\tau}$ then the contribution indicates that token $\tau$ is more typical of text $T_1$ than of text $T_2$, and vice versa. Such decompositions may be visualized as in Fig. 1.

\begin{center}
\includegraphics[width=4.31in,height=0.75\textheight,keepaspectratio]{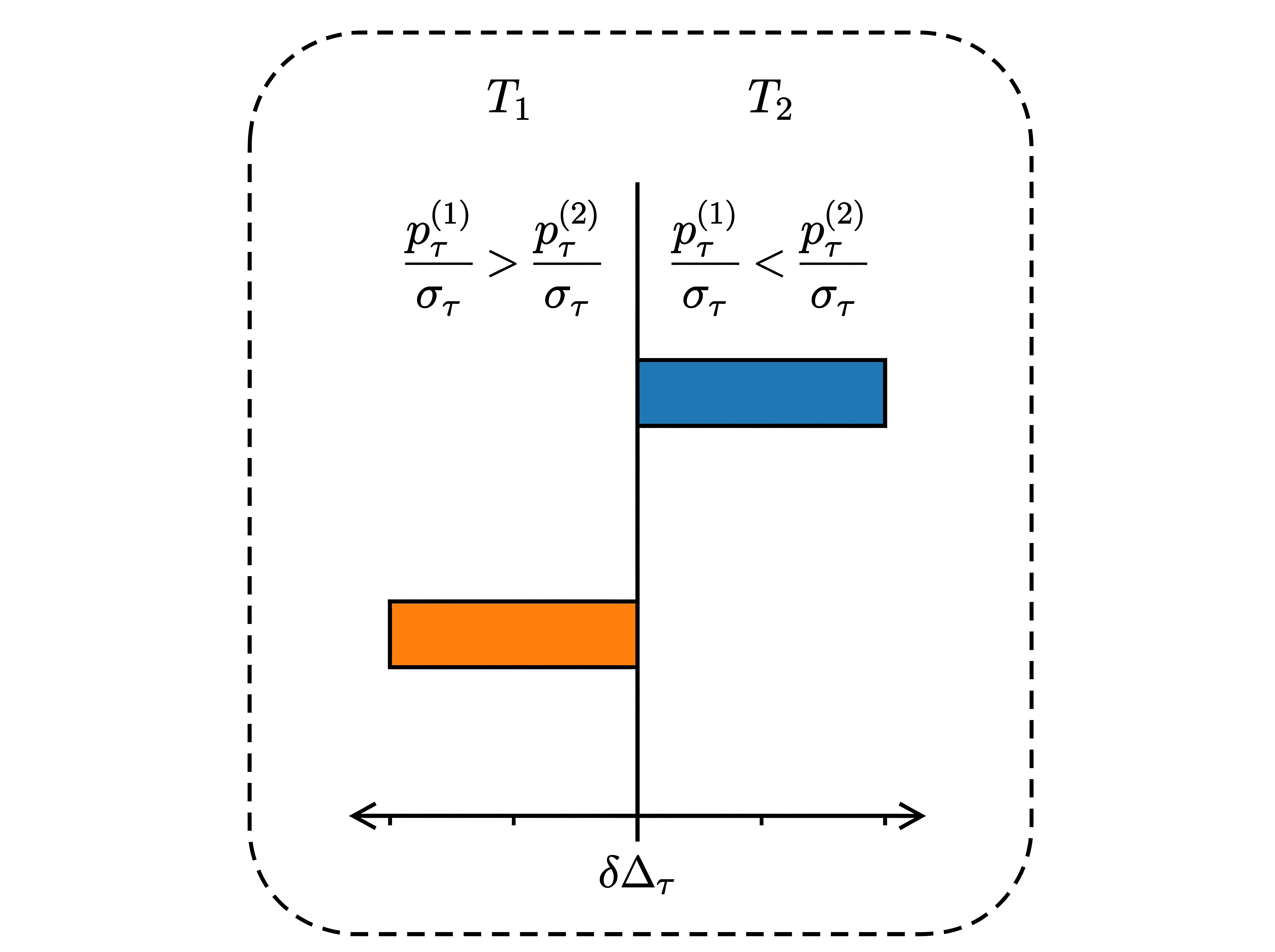}\par
\noindent Fig. 1 Visualization of tokenlevel contributions to Delta.
\end{center}

It is important to distinguish this decomposition from a simple comparison of raw or standardized frequencies. The contribution of a token is determined not only by its individual frequency difference but also by the geometry of the chosen distance function. In this sense, token-level contributions reflect the structure of the metric itself.

\section*{4.1 Cosine Delta}

Assuming that the z-vectors have been L$_2$-normalized, the overall metric for Cosine Delta takes the form:

\begin{equation}\tag{13}
\Delta^{\cos}(T_1,T_2)=1-\sum_{i=1}^{n}z_i^{(1)}z_i^{(2)}
\end{equation}

Leaving aside the additive constant 1 (which does not affect comparisons within a fixed corpus), the contribution of token $\tau$ can be written as:

\begin{equation}\tag{14}
\delta\Delta_\tau^{\cos}=-z_\tau^{(1)}z_\tau^{(2)}
\end{equation}

Unlike Burrows’s Delta, these contributions arise from products rather than sums and may thus be either positive or negative. Four scenarios can be distinguished for the pair $z_\tau^{(1)},z_\tau^{(2)}$ (Table 1)

\begin{samepage}
\noindent Table 1. Interpretation of the four possible sign configurations of standardized token pairs in Cosine Delta.
\begin{center}\begin{tabular}{ccc p{3.5in}}\hline
$z_\tau^{(1)}$ & $z_\tau^{(2)}$ & $\delta\Delta_\tau^{\cos}$ & Interpretation\\\hline
$+$ & $+$ & $-$ & Both components exceed their corpus means\\
$+$ & $-$ & $+$ & The first exceeds while the second falls below\\
$-$ & $+$ & $+$ & The second exceeds while the first falls below\\
$-$ & $-$ & $-$ & Both fall below their corpus means\\\hline
\end{tabular}\end{center}
\end{samepage}

Tokens that yield $\delta\Delta_\tau^{\cos}>0$ increase the stylistic distance, while those with $\delta\Delta_\tau^{\cos}<0$ reduce it. The overall Cosine Delta reflects which group predominates and by how much. To identify the most influential tokens for text difference we select those with the largest positive $\delta\Delta_\tau^{\cos}$ and visualizes their contributions analogously to Figure 1.

\section*{4.2 Jensen–Shannon Delta}

Token contributions for Jensen–Shannon Delta are extracted in the same manner as for canonical Burrows’s Delta in equation:

\begin{equation}\tag{15}
\begin{aligned}\delta JSD_\tau={}&m_\tau\log_2\frac1{m_\tau}\\&-\left(\pi_1\rho_\tau^{(1)}\log_2\frac1{\rho_\tau^{(1)}}+\pi_2\rho_\tau^{(2)}\log_2\frac1{\rho_\tau^{(2)}}\right)\end{aligned}
\end{equation}

where $m_\tau$ is the probability of token $\tau$ in the hybrid distribution $M$.

\section*{4.3 Rank-Turbulence Delta}

For Rank-Turbulence divergence the token contribution is analogous to that for canonical Burrows’s Delta and may be written as:

\begin{equation}\tag{16}
\delta RTD_\tau=\frac1{\mathcal N_{1,2;\alpha}}\frac{\alpha+1}{\alpha}\left|\frac1{(r_\tau^{(1)})^\alpha}-\frac1{(r_\tau^{(2)})^\alpha}\right|^{1/(\alpha+1)}
\end{equation}

where $r_\tau^{(1)}$ and $r_\tau^{(2)}$ are the ranks of token $\tau$ in texts $T_1$ and $T_2$, respectively, $\alpha$ is the tuning parameter, and $\mathcal N_{1,2;\alpha}$ is the normalization constant defined in Equation (11).

\section*{5. Experiments on Clustering and Classification}

The experiments reported below are designed to evaluate the behaviour and robustness of distance-based measures under exploratory stylometric conditions. They are not intended to compete with supervised state-of-the-art authorship attribution pipelines, but rather to assess the internal consistency and comparative performance of Delta-like metrics across varying corpus structures.

As a first experiment, we performed clustering on a database consisting of three groups of novels written in English, French, and German. Each group contained three novels by twenty-five different authors, giving a total of seventy-five texts per language. The Partitioning Around Medoids algorithm was employed for clustering, and clustering quality was assessed with the adjusted Rand index, computed for various numbers of most frequent words (mfw), as shown in Fig. 2. Such an experimental design permits direct comparison with approaches previously reported in the authorship-attribution literature, notably Jannidis and Evert (Jannidis et al. 2015; Evert et al. 2017).

\begin{center}
\includegraphics[width=\textwidth,height=0.75\textheight,keepaspectratio]{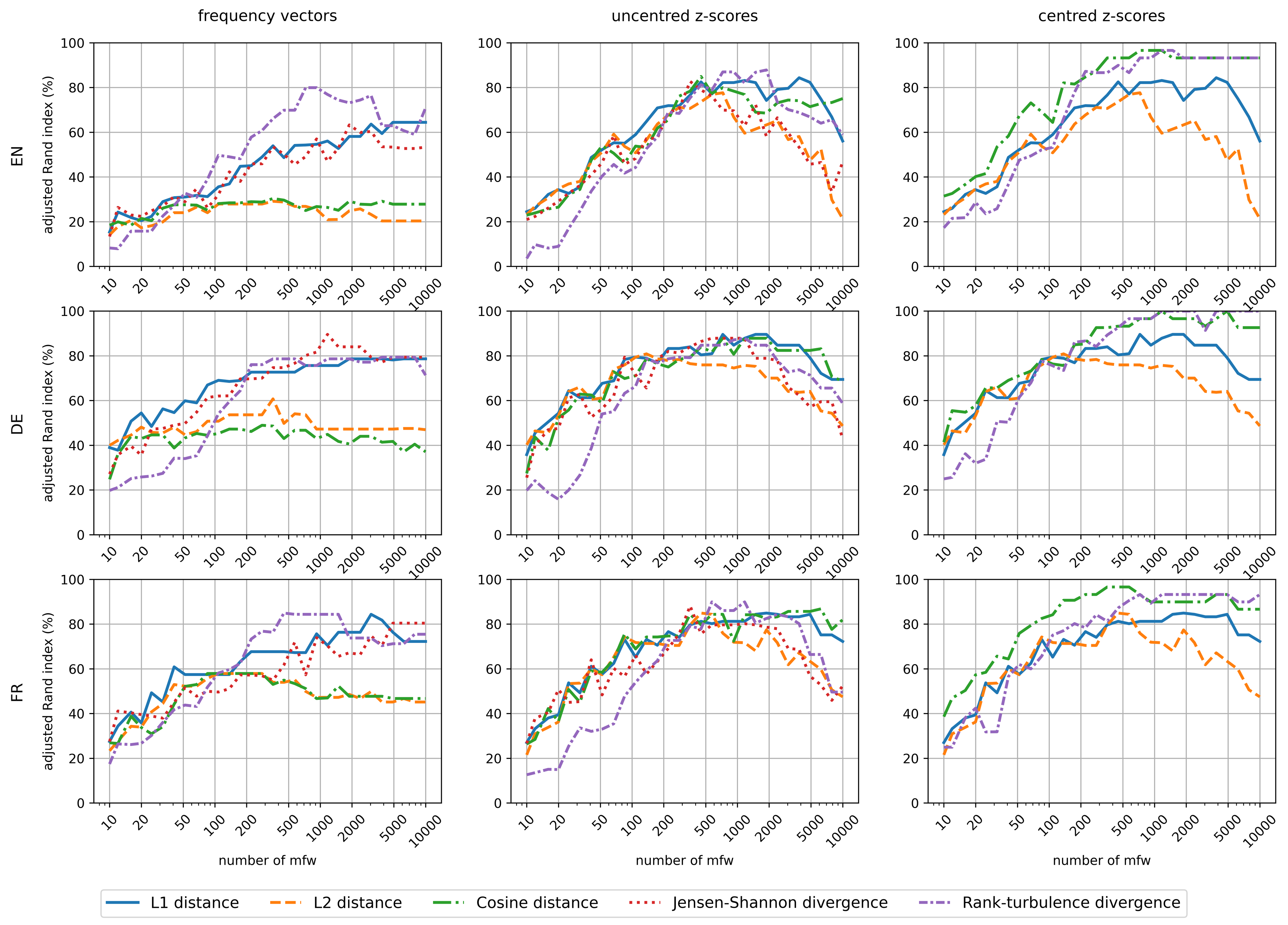}\par
\noindent Fig. 2 Clustering quality of the different measures in the authorship-attribution task as a function of mfw and language. L$_1$ and L$_2$ distances correspond to Burrows’s Delta and Quadratic (Euclidean) Delta when centred and uncentred z-scores are used, respectively. Cosine distance based on centred z-scores corresponds to Cosine Delta)
\end{center}

Rank-Turbulence and Cosine Delta demonstrate consistently high performance across languages and mfw ranges, suggesting that both metrics provide stable representations of stylistic structure under exploratory conditions. It is noteworthy that Rank-Turbulence maintains relatively high performance both with regular z-scores and with raw relative-frequency vectors, which are generally ineffective for authorship attribution. Jensen–Shannon Delta was not applied to centred z-scores because it presupposes non-negative input values.

Because Rank-Turbulence Delta includes the parameter, we measured clustering quality for a range of $\alpha$ values (Fig. 3).

\begin{center}
\includegraphics[width=\textwidth,height=0.75\textheight,keepaspectratio]{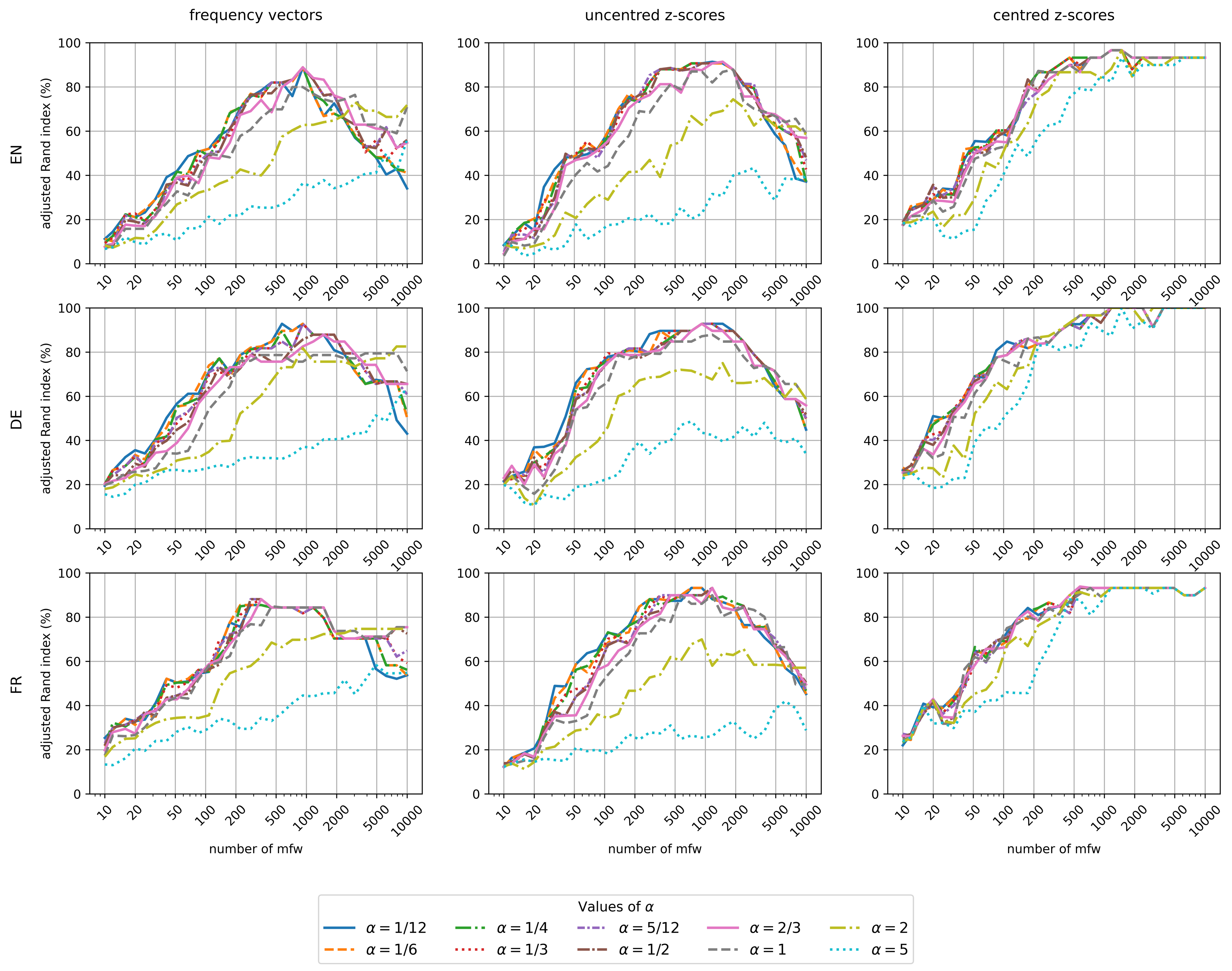}\par
\noindent Fig. 3 Clustering quality of Rank-Turbulence Delta for different $\alpha$ values in the authorship-attribution task as a function of mfw and language.
\end{center}

Graphical analysis indicates that clustering quality is maximized at lower values of $\alpha$, while larger values lead to a noticeable decline in accuracy. This pattern is particularly pronounced when the metric is applied to non-centred representations, where all components are non-negative. In this configuration, increasing $\alpha$ shifts analytical emphasis toward the highest-frequency tokens; beyond a certain threshold, this concentration reduces discriminative power and leads to performance degradation.

Across languages, the strongest and most stable performance is observed at relatively small or intermediate $\alpha$ values. This suggests that mid-frequency words, rather than exclusively dominant lexical items, carry substantial stylistic information in attribution tasks. On average, Rank-Turbulence performs best with mfw values between 500 and 2,000, where the balance between structural vocabulary and more distinctive lexical signals appears most effective.

To this end, the methods were tested on the Russian-language SOCIOLIT corpus of literary works, which contains 639 texts by 89 authors written from the eighteenth to the twenty-first centuries and is characterized by substantial temporal and stylistic variation. Before constructing z-vectors, the texts were lemmatized with the pymorphy3 library, thereby minimizing the influence of morphological variation on the distance measures (see Appendix A for more details).

\begin{center}
\includegraphics[width=\textwidth,height=0.75\textheight,keepaspectratio]{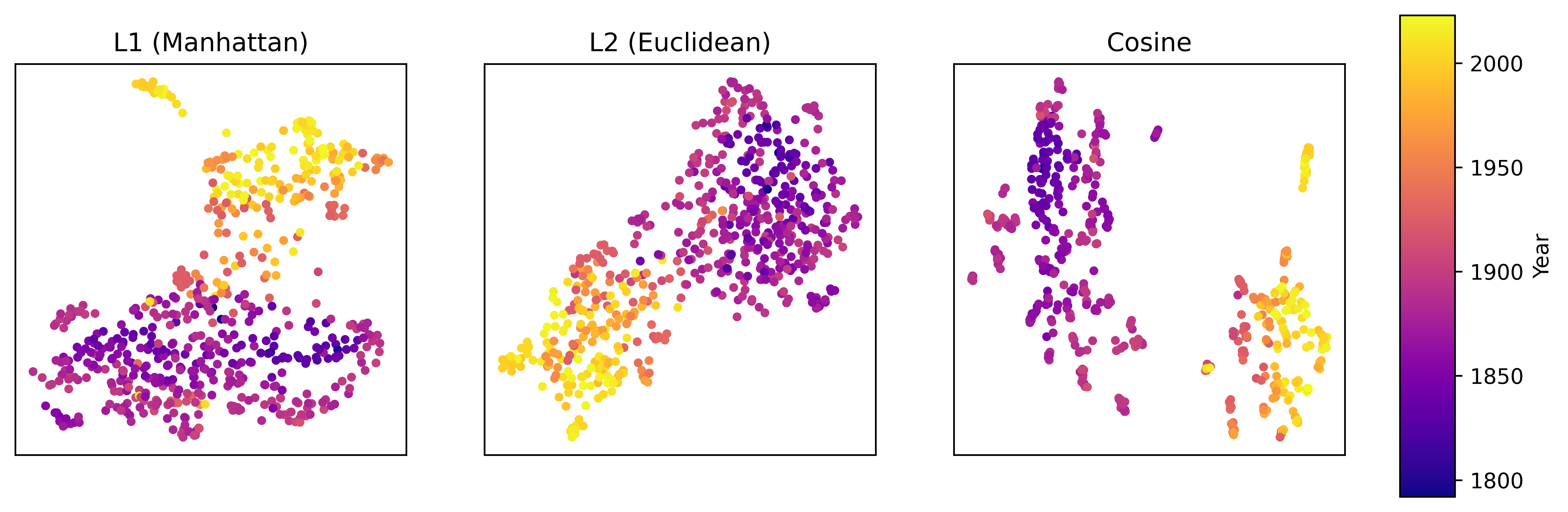}\par
\noindent Fig. 4 Z-vectors of the corpus texts after dimensionality reduction with UMAP using different values of the argument metric.
\end{center}

For visualizing corpus structure the UMAP dimensionality-reduction method (McInnes L., Healy J. and Melville J. 2018) was applied with Manhattan, Euclidean, and Cosine metrics – that is, with canonical Burrows’s Delta, Quadratic Delta, and Cosine Delta as the underlying text distances. Visual inspection showed a pronounced tendency for texts to cluster according to their time of composition (Fig. 4)

Authorship attribution was performed with a nearest-neighbour method in which distances between texts were computed using the measures under investigation. Classification accuracy was evaluated by Leave-One-Out Cross-Validation (LOOCV). This procedure mirrors the real-world attribution scenario, where the authorship of a given work is decided on the basis of its similarity to other texts with known provenance.

The principal quality criterion was the balanced accuracy score, selected because the distribution of works across authors is uneven, which complicates the classification task and better reflects real-world conditions. Balanced accuracy makes it possible to assess attribution quality correctly when the number of works per author varies widely, as is typical of literary corpora.

\begin{center}
\includegraphics[width=\textwidth,height=0.75\textheight,keepaspectratio]{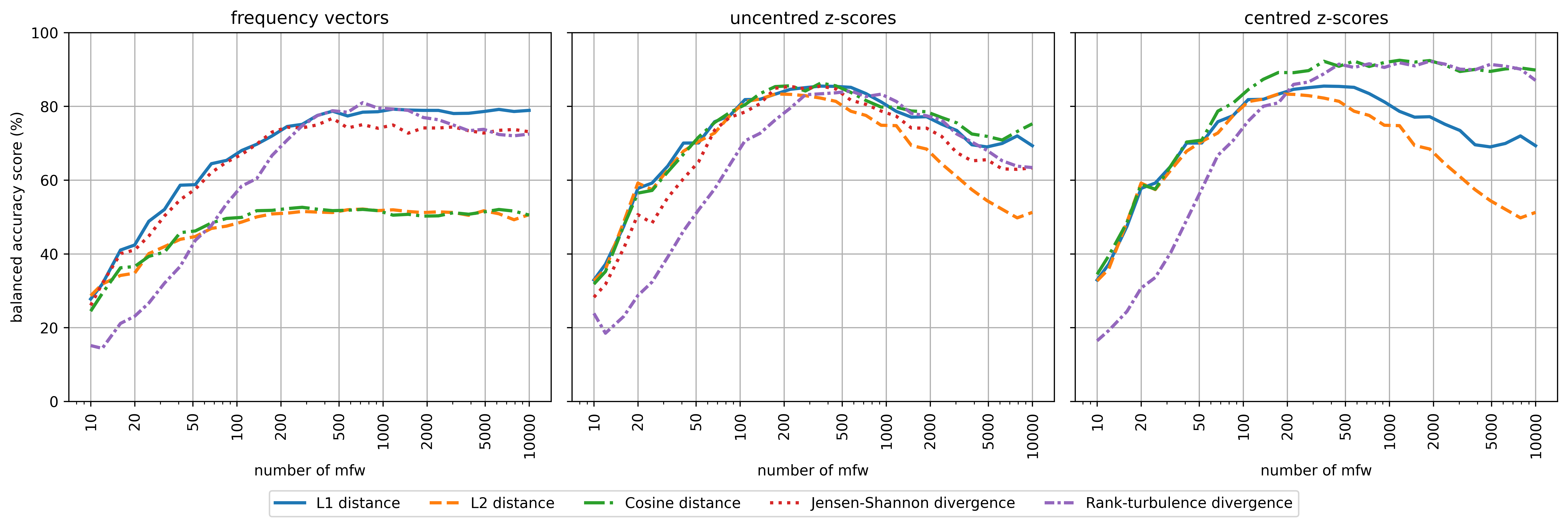}\par
\noindent Fig. 5 Attribution quality of different Delta measures as a function of the mfw considered.
\end{center}

The higher performance results are again obtained with Cosine Delta and Rank-Turbulence Delta, which remain stable across an mfw range of 500-2,000 (Fig. 5). A further observation deserves particular attention. Rank-Turbulence Delta demonstrates comparatively strong performance even when applied to centred z-scores. In this representation, tokens with large negative z-values – that is, words used substantially less frequently by a given author than by the corpus average – are placed at the lower end of the rank distribution. Although such tokens encode meaningful stylistic absence, they exert limited influence on the rank-based divergence.

This asymmetry suggests that, within a rank framework, the reorganization of prominent lexical signals may be more structurally informative than the relative scarcity of particular items. In other words, words that prevail in an author’s usage appear to contribute more decisively to stylistic separation than words that are merely underrepresented.

\section*{6. Experiments on Interpretation}

The interpretation experiments build on the token-level decomposition introduced in Section 4. As noted above, the canonical Burrows’s Delta and its modifications enable the researcher not only to calculate distances between texts but also to reveal the individual token contributions that give rise to those distances. This capability permits an interpretable stylometric examination that identifies the words most characteristic of a particular author.

To illustrate the approach we visualize the contributions of individual words to the distances between the writings of F. M. Dostoevsky (orange) and L. N. Tolstoy (blue). First, all z-vectors of texts belonging to the two authors are extracted from the corpus. We then compute a mean z-vector for each author, representing the average frequency distribution of words in his works, and finally calculate the distances between those mean vectors while retaining the contribution of every single word (Fig. 6).

\begin{center}
\includegraphics[width=\textwidth,height=0.75\textheight,keepaspectratio]{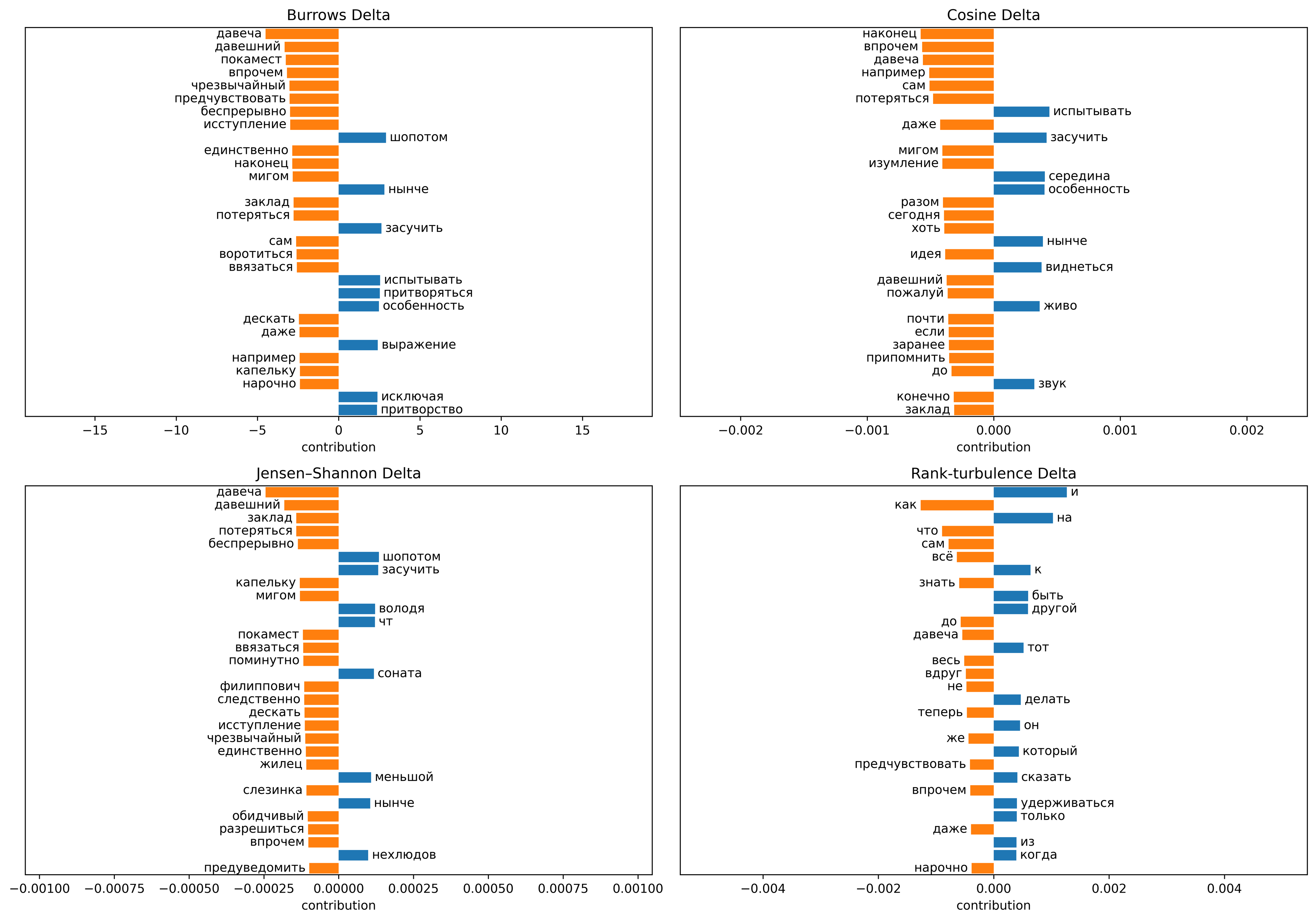}\par
\noindent Fig. 6 Tokens making the largest contributions to Burrows’s Delta, Cosine Delta Jensen–Shannon Divergence and Rank-Turbulence Divergence ($\alpha=1$) between the works of Dostoevsky (orange) and Tolstoy (blue).
\end{center}

In the visualizations, the words exerting the greatest influence on the distances are highlighted. The detailed behaviour of contribution weights across varying $\alpha$ values for Rank-Turbulence Delta is presented in Appendix В. The figures show that small and intermediate $\alpha$ values produce relatively diverse and stable sets of high-contributing tokens. As $\alpha$ increases, the contribution mass gradually concentrates on the most frequent functional items, resulting in reduced lexical diversity and weaker discriminative nuance. At lower $\alpha$ values, the sets of high-contributing tokens largely resemble those identified by other distance measures.

To ensure that the identified top-contributing tokens are not artefacts of a particular parameter choice or document sample, we performed two robustness checks: (i) variation of mfw size, and (ii) document-level bootstrap resampling (see Appendix C). Across metrics, the mean Jaccard overlap of the top-30 tokens under bootstrap resampling indicating substantial structural stability. Additionally, removing the top-contributing tokens consistently reduced inter-author distances in the expected direction across Burrows’s Delta, Cosine Delta, Jensen-Shannon Delta, and Rank-Turbulence Delta, confirming their functional role in document separation.

It is important to remember that, under the proposed procedure, the final result is affected not only by the two authors under investigation but also by the remaining texts in the corpus, because they take part in the standardization process. For that reason it may be useful to restrict the corpus to the works of the authors being compared, or to texts belonging to a single period or genre; such a strategy can better reflect the specific features of each author’s idiolect.

Tokens need not be single words: employing three- to five-gram word tokens substantially enhances the value of the visualization by revealing stable word combinations characteristic of an author rather than isolated lexemes. The method thus yields a more informative picture of stylistic differences (Fig. 7).

\begin{center}
\includegraphics[width=\textwidth,height=0.75\textheight,keepaspectratio]{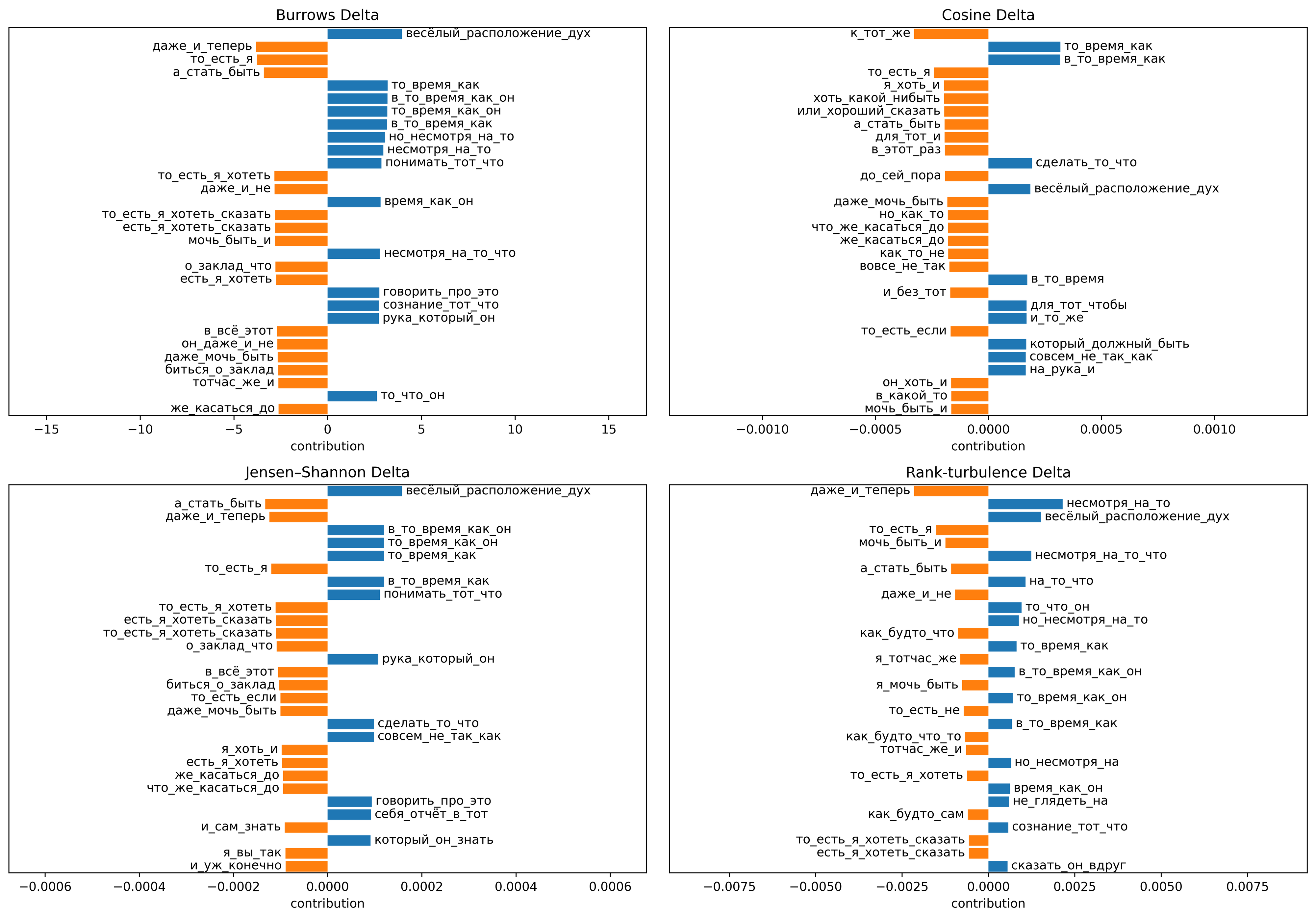}\par
\noindent Fig. 7 Token contributions (3–5-grams) to Burrows’s Delta, Cosine Delta, Jensen–Shannon Divergence and Rank-Turbulence Divergence.
\end{center}

For example, Tolstoy is marked by phrases such as «лет тому назад» (‘years ago’), «в расположении духа» (‘in a frame of mind’), and «стать ходить по комнате» (‘to begin pacing the room’), whereas Dostoevsky typically employs pronominal adverbs and adjectives like «какой-то» (‘some sort of’), «как-то» (‘somehow’) and the word «особенно» (‘especially’) in combination with them.

\section*{7. Discussion}

The present study repositions Delta as a flexible family of distance measures suitable for exploratory stylometry. In contexts such as clustering, network visualization, diachronic comparison, and corpus exploration, interpretability, stability, and theoretical transparency are often as important as raw classification accuracy. Within this framework, the extension of Delta into probabilistic and rank-based geometries broadens its analytical scope while preserving its foundational intuition: stylistic distance arises from systematic contrasts in lexical distributions.

Reinterpreting z-vectors as probability or rank distributions does not alter the underlying feature space, but changes the geometry of comparison. This shift allows the use of divergence measures such as Jensen-Shannon and Rank-Turbulence, which provide alternative perspectives on stylistic contrast. Importantly, the results demonstrate that these extensions remain compatible with the established logic of Delta rather than replacing it.

The experiments further suggest that mid-frequency words play a particularly important role in stylistic discrimination. Both clustering performance and token-level analyses indicate that moderate sensitivity settings – especially in Rank-Turbulence – yield the most stable and informative results. When $\alpha$ becomes large, attention shifts disproportionately toward high-frequency function words, reducing lexical diversity and interpretative nuance. This finding reinforces a long-standing observation in stylometry: neither exclusively dominant nor exclusively rare items suffice to capture stylistic identity; rather, structurally embedded mid-frequency vocabulary carries substantial signal.

At the same time, the rank-based perspective reveals an asymmetry between lexical presence and lexical absence. In centred representations, strongly negative deviations tend to occupy lower positions in the rank hierarchy and therefore contribute less prominently to divergence. Although this reduces the relative weight of negatively marked tokens, clustering quality does not deteriorate; on the contrary, the resulting solutions often appear more stable across parameter settings. This suggests that stylistic differentiation within a rank framework is driven more by prominent lexical signals than by systematic underrepresentation. The theoretical implications of this asymmetry, particularly for the study of negative stylistic markers, merit further investigation.

Several limitations should be acknowledged. Parameter selection, especially for $\alpha$, remains context-dependent and may require corpus-specific stability analysis. Moreover, although the present experiments cover multiple languages and corpora, further testing on heterogeneous historical datasets would strengthen the generalizability of the findings. Importantly, the application of Rank-Turbulence divergence requires transforming frequency or z-score vectors into rank distributions. This transformation entails an explicit reordering of features, with asymptotic complexity comparable to sorting operations. While such costs are moderate for standard stylometric vocabularies, they may become non-negligible for very large corpora or in settings involving extensive pairwise comparisons. Finally, the interpretative framework proposed here does not replace close reading; rather, it provides a structured quantitative map of lexical contrasts that may guide subsequent qualitative analysis.

In this sense, the study contributes to the ongoing effort to integrate statistical modelling with philological reasoning, demonstrating that refinements of classical distance measures can enhance both analytical flexibility and interpretative clarity within exploratory stylometry.

\section*{Data and Code Availability}

Code and data used in this study are publicly available at:

\url{https://github.com/DDPronin/Rank-Turbulence-Delta}

Due to copyright restrictions, full-text data from the SOCIOLIT corpus are not redistributed. However, metadata and derived representations are provided in the repository, along with instructions for reproducing the experiments.

\section*{References}
\begin{sloppypar}

Argamon, S. (2008) ‘Interpreting Burrows’s Delta: Geometric and Probabilistic Foundations’, Literary and Linguistic Computing, 23/2: 131–47. doi:10.1093/llc/fqn003.

Beausang, C. (2022) ‘Diachronic Delta: A Computational Method for Analysing Periods of Accelerated Change in Literary Datasets’, Digital Scholarship in the Humanities, 37/3: 644–59. doi:10.1093/llc/fqab041.

Bevendorff, J. et al. (2021) ‘Overview of PAN 2021: Authorship Verification, Profiling Hate Speech Spreaders on Twitter, and Style Change Detection’, in CLEF 2021 Labs and Workshops: Notebook Papers. CEUR-WS.org. Available at: http://ceur-ws.org/ (accessed 26 February 2026)

Burrows, J. (2002) ‘“Delta”: A Measure of Stylistic Difference and a Guide to Likely Authorship’, Literary and Linguistic Computing, 17/3: 267–87. doi:10.1093/llc/17.3.267.

Cafiero, F. and Camps, J.-B. (2019) ‘Why Molière Most Likely Did Write His Plays’, Science Advances, 5: eaax5489. doi:10.1126/sciadv.aax5489.

Cafiero, F. and Camps, J.-B. (2023) ‘Who Could Be behind QAnon? Authorship Attribution with Supervised Machine-Learning’, Digital Scholarship in the Humanities, 38/4: 1418–30. doi:10.1093/llc/fqad061.

Camps, J.-B., Clérice, T. and Pinche, A. (2021) ‘Noisy Medieval Data, from Digitized Manuscript to Stylometric Analysis: Evaluating Paul Meyer’s Hagiographic Hypothesis’, Digital Scholarship in the Humanities, 36/suppl. 2: ii49–ii71. doi:10.1093/llc/fqab033.

Dodds, P. S. et al. (2023) ‘Allotaxonometry and Rank-Turbulence Divergence: A Universal Instrument for Comparing Complex Systems’, EPJ Data Science, 12/1: 37. doi:10.1140/epjds/s13688-023-00386-7.

Dviniatin, F. N. and Kovalev, B. V. (2024) “Sirin is not inferior to Leonov”: Burrows’s Delta for Stylometric Analysis of Russian Novels of the Interwar Period’, Shagi/Steps, 10/2: 207–29. doi:10.22394/2412-9410-2024-10-2-207-229.

Eder M., Rybicki J., Kestemont M. (2016) ‘Stylometry with R: A package for computational text analysis’, The R Journal,. 8/1: 107-21 Available at: https://journal.r-project.org/articles/RJ-2016-007/RJ-2016-007.pdf (accessed 20 February 2026).

Eder, M. (2016) ‘Rolling Stylometry’, Digital Scholarship in the Humanities, 31/3: 457–69. doi:10.1093/llc/fqv010.

Eder, M. (2017) ‘Visualization in Stylometry: Cluster Analysis Using Networks’, Digital Scholarship in the Humanities, 32/1: 50–64. doi:10.1093/llc/fqv061.

Evert, S. et al. (2017) ‘Understanding and Explaining Delta Measures for Authorship Attribution’, Digital Scholarship in the Humanities, 32/suppl. 2: ii4–ii16. doi:10.1093/llc/fqx023.

Gallagher, R. J. et al. (2021) ‘Generalised Word-Shift Graphs: A Method for Visualising and Explaining Pairwise Comparisons between Texts’, EPJ Data Science, 10/1: 4. doi:10.1140/epjds/s13688-021-00264-3.

Hoover, D. L. (2004) ‘Testing Burrows’s Delta’, Literary and Linguistic Computing, 19/4: 453–75. doi:10.1093/llc/19.4.453.

Jannidis, F. et al. (2015) ‘Improving Burrows’s Delta–An Empirical Evaluation of Text Distance Measures’, in Proceedings of the Digital Humanities 2015 Conference, 11: 10.

Kestemont, M. et al. (2019) ‘Overview of the Cross-Domain Authorship Attribution Task at PAN 2019’, in CLEF 2019 Labs and Workshops: Notebook Papers. CEUR-WS.org. Available at: http://ceur-ws.org/Vol-2380/paper\_264.pdf (accessed 20 February 2026).

Kestemont, M. et al. (2021) ‘Overview of the Cross-Domain Authorship Verification Task at PAN 2021’, in CLEF 2021 Labs and Workshops: Notebook Papers. CEUR-WS.org. Available at: http://ceur-ws.org/Vol-2936/paper-147.pdf (accessed 20 February 2026).

Kocher, M. and Savoy, J. (2019) ‘Evaluation of Text Representation Schemes and Distance Measures for Authorship Linking’, Digital Scholarship in the Humanities, 34/1: 189–207. doi:10.1093/llc/fqy013.

Kovalev, B. V. (2024) ‘From Classics to Digital Philology: on the Origin and Growth of Stylometry’, Philologia Classica, 19/2: 347–60. doi:10.21638/spbu20.2024.211.

Kovalev, B. V., and Dviniatin, F. N. (2025). The Burrows’ Delta: towards a linguistic interpretation. Acta Linguistica Petropolitana, 21(3): 263–96. doi:10.30842/alp23065737213

McInnes L., Healy J. and Melville J. (2018) ‘UMAP: Uniform Manifold Approximation and Projection’, Journal of Open Source Software, 3/29: 861. doi:10.21105/joss.00861.

Nilsson-Fernàndez, P. (2025) ‘Who Wrote Els Quaderns d’en Marc? A Stylometric Inquiry into Twentieth-Century Catalan Popular Literature’, Digital Scholarship in the Humanities. doi:10.1093/llc/fqaf096.

Orekhov, B. V. (2021) ‘Text and Translation by Vladimir Nabokov through the Prism of Stylometry’, Novyi filologicheskii vestnik, 58/3: 200–13. doi:10.54770/20729316\_2021\_3\_200.

Päpcke, S. et al. (2023) ‘Stylometric Similarity in Literary Corpora: Non-Authorship Clustering and Deutscher Novellenschatz’, Digital Scholarship in the Humanities, 38/1: 277–95. doi:10.1093/llc/fqac039.

Proisl, T. et al. (2018) ‘Delta vs. N-Gram Tracing: Evaluating the Robustness of Authorship Attribution Methods’, in Proceedings of LREC 2018.

Stamatatos, E. (2009) ‘A Survey of Modern Authorship Attribution Methods’, Journal of the American Society for Information Science and Technology, 60/3: 538–56. doi:10.1002/asi.21001.

Velikanova, N. P. and Orekhov, B. V. (2019) ‘Digital Textual Criticism: Text Attribution on the Example of the Novel by M. A. Sholokhov And Quiet Flows the Don’, Mir Sholokhova, 1: 70–82.

SOCIOLIT Corpus (n.d.) SOCIOLIT Corpus. Available at https://sociolit.ru (accessed 14 July 2025)

\end{sloppypar}
\clearpage
\section*{Appendix A: Experiment Details}

The term–frequency matrices and author labels for the non-Russian datasets were obtained from the public repository accompanying Evert et al. (2017): https://github.com/schtepf/ExplainingDelta

The repository is distributed under an open licence permitting reuse for research purposes. Using the same curated material ensures direct comparability with previous studies on Delta measures conducted on this benchmark corpus. No modifications to the corpus composition were introduced beyond the preprocessing steps described below.

Russian-language texts were downloaded from the SOCIOLIT corpus (https://sociolit.ru). Some texts in this corpus, particularly those by contemporary authors, are subject to copyright restrictions and are available only via restricted internal access.

For transparency and reproducibility, the full list of texts used in the Russian experiments is provided in the project repository (see link below). Specifically, metadata for all Russian texts included in the experiments are available in: data/russian/russian\_metadata.xlsx .Term-frequency matrices derived from these texts are also included in the repository.

The preprocessing pipeline was designed to ensure consistency across languages while preserving stylistically relevant information. The following steps were applied:

1. Only Cyrillic and Latin alphabetic characters were retained; all other characters were removed.

2. All texts were converted to lowercase.

3. For Russian texts, lemmatisation was performed using pymorphy3.

4. Stopwords were not removed.

5. No manual filtering of high-frequency functional words was applied.

Term-frequency (TF) matrices were constructed using the 20,000 most frequent word types in each corpus. All experiments reported in the article were conducted using the first 10,000 most frequent words (mfw) drawn from this global frequency ranking. Subsequently, both centred and uncentred z-standardised representations were computed for each TF matrix.

From the resulting matrices, the following procedures were applied:

1. Computation of centred and uncentred z-scores.

2. Calculation of the distance measures described in the article (Classic Delta, Cosine Delta, Jensen–Shannon Delta, Rank-Turbulence Delta).

3. Clustering and classification experiments as detailed in Sections 5 and 6.

To ensure numerical stability in Jensen–Shannon divergence, Laplace smoothing was applied by adding a small (1e-10) value prior to normalization.

A detailed implementation of the full preprocessing and experimental pipeline is available in the project repository.

\clearpage
\section*{Appendix B. Sensitivity of Rank-Turbulence token contributions to alpha parameter}

Figure B1 illustrates the evolution of token-level contributions in Rank-Turbulence Delta across a range of $\alpha$ values (from 1/12 to 5). Each panel displays the top contributing tokens for a fixed number of most frequent words.

For small $\alpha$ values, contributions are distributed across a broader spectrum of lexical items, including mid-frequency words. As $\alpha$ increases, the contribution profile becomes increasingly dominated by high-frequency function words, reflecting the theoretical property of the metric: large $\alpha$ values amplify sensitivity to top-ranked tokens. This visualisation supports the claim that moderate $\alpha$ values provide the most balanced interpretative resolution.

\clearpage
\begin{landscape}
\begin{center}
\includegraphics[width=0.95\linewidth,height=0.8\textheight,keepaspectratio]{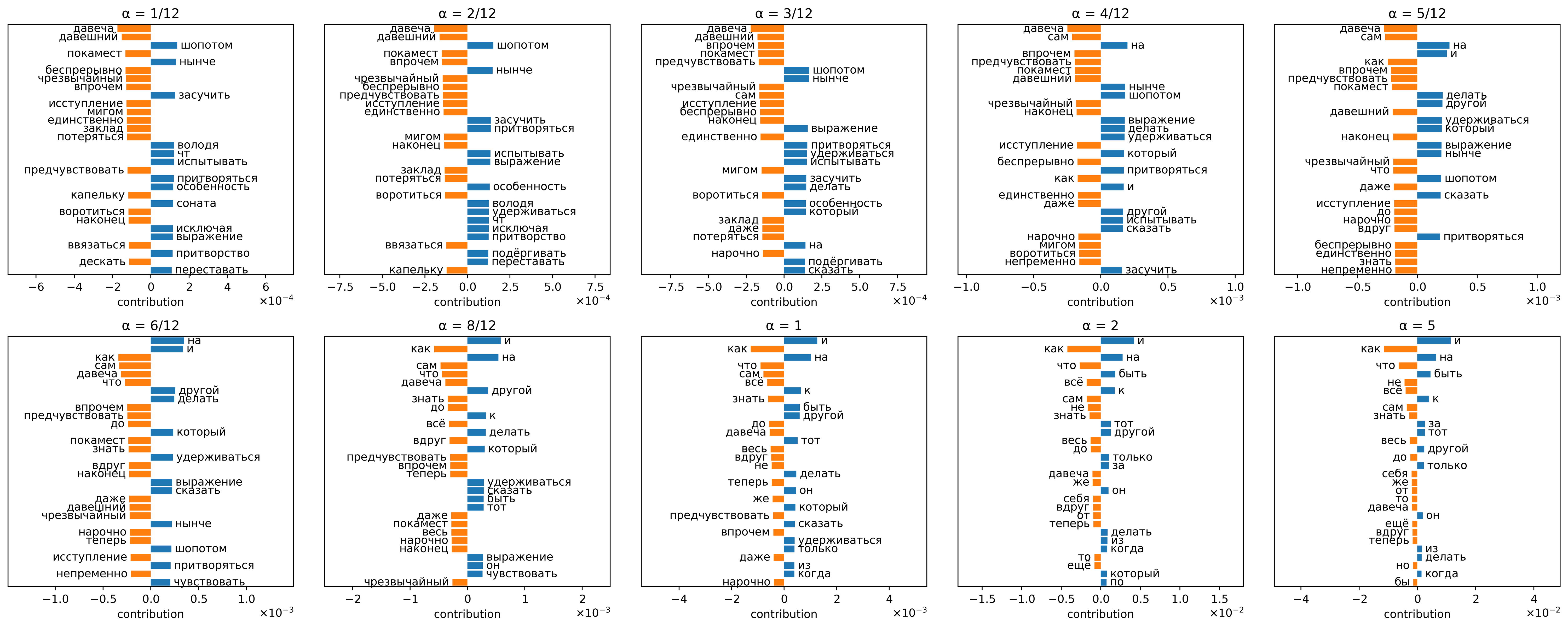}\par
\noindent Fig. B1 Evolution of token-level contributions in Rank-Turbulence Delta across a range of parameter values
\end{center}
\end{landscape}

\clearpage
\section*{Appendix C. Robustness of Top-Contributing Tokens}

To assess whether the identified top-contributing tokens represent structurally meaningful stylistic markers rather than artefacts of parameter choice or sampling variation, we conducted two robustness checks: (i) variation of the mwf size and (ii) document-level bootstrap resampling. All experiments were performed on the Russian corpus for Tolstoy-Dostoevsky author pair (31 texts). The same preprocessing pipeline was applied as described in the Appendix A. Term-frequency matrices were constructed using the N most frequent words (mfw), with N varied in controlled perturbations. For contribution-based analyses, the top-K tokens were defined as those with the largest absolute contribution to the corresponding distance metric. Unless otherwise stated, K was fixed to 30. Bootstrap experiments were performed by resampling documents with replacement within each author, and recomputing token contributions for each resampled dataset.

To evaluate sensitivity to the choice of vocabulary size, we computed token contributions under neighbouring mfw values around 3,000 (e.g.  2,800, 3,000, and 3,200). For each configuration, the top-K tokens were extracted and compared using Jaccard overlap. Across all metrics (Burrows’s Delta, Cosine Delta, Jensen–Shannon Divergence, and Rank-Turbulence Divergence), the Jaccard overlap between adjacent mwf settings remained consistently substantial. This indicates that moderate changes in vocabulary size do not qualitatively alter the identity of the most influential tokens. These results suggest that the interpretative conclusions drawn from top-contributing tokens are not dependent on a narrowly tuned mfw parameter but reflect a stable structural property of the authorial contrast (Table C1–C4).

\begin{samepage}
\noindent Table C1. Stability of top-contributing tokens under variation of mwf size for Burrows’s Delta.
\begin{center}\begin{tabular}{ll}\hline
mfw & Jaccard Overlap\\
2,800 & 0.935\\
3,000 & 1.000 (base)\\
3,200 & 0.875\\
\hline\end{tabular}\end{center}
\end{samepage}

\begin{samepage}
\noindent Table C2. Stability of top-contributing tokens under variation of mwf size for Cosine Delta.
\begin{center}\begin{tabular}{ll}\hline
mfw & Jaccard Overlap\\
2,800 & 0.935\\
3,000 & 1.000 (base)\\
3,200 & 0.935\\
\hline\end{tabular}\end{center}
\end{samepage}

\begin{samepage}
\noindent Table C3. Stability of top-contributing tokens under variation of mwf size for Jensen–Shannon Divergence.
\begin{center}\begin{tabular}{ll}\hline
mfw & Jaccard Overlap\\
2,800 & 0.818\\
3,000 & 1.000 (base)\\
3,200 & 0.765\\
\hline\end{tabular}\end{center}
\end{samepage}

\begin{samepage}
\noindent Table C4. Stability of top-contributing tokens under variation of mwf size for Rank-Turbulence Divergence.
\begin{center}\begin{tabular}{ll}\hline
mfw & Jaccard Overlap\\
2,800 & 0.875\\
3,000 & 1.000 (base)\\
3,200 & 0.875\\
\hline\end{tabular}\end{center}
\end{samepage}

To further assess robustness under sampling variation, we performed document-level bootstrap resampling. For each resample, token contributions were recomputed and the resulting top-K lists were compared to those obtained from the full dataset. The mean Jaccard overlap of the top-30 tokens across bootstrap iterations ranged between approximately 0.37 and 0.50, depending on the metric, with standard deviations between 0.07 and 0.10. Given the relatively small number of documents per author, this level of overlap indicates substantial structural stability. In small samples, individual texts can meaningfully influence frequency distributions; therefore, perfect invariance would not be expected. Instead, the observed overlap values demonstrate that a core subset of highly discriminative tokens persists under document resampling, supporting the interpretability of the contribution analysis. Among the metrics, Cosine Delta and Jensen-Shannon Divergence exhibited slightly higher overlap values, whereas Rank-Turbulence Divergence showed lower overlap, consistent with its sensitivity to rank perturbations. Nevertheless, in all cases, the persistence of approximately 40–50\% of the top contributors indicates a stable signal rather than sampling noise (Table C5)

\begin{samepage}
\noindent Table C5. Bootstrap stability of top-30 token contributions across distance metrics.
\begin{center}\begin{tabular}{lll}\hline
Metric & Mean Jaccard & Std. Dev.\\
Burrows's Delta & 0.412 & 0.074\\
Cosine Delta & 0.495 & 0.093\\
Jensen–Shannon Divergence & 0.460 & 0.102\\
Rank-Turbulence Divergence & 0.370 & 0.075\\
\hline\end{tabular}\end{center}
\end{samepage}

Finally, to confirm that the identified tokens are functionally responsible for document separation, we conducted a removal experiment. The top-K contributing tokens were set to zero in both author profiles, and inter-author distances were recomputed. Across all four metrics, removing the top-contributing tokens led to a consistent reduction in inter-author distance. Moreover, the magnitude of reduction increased monotonically with K (e.g. K = 10, 50, 100), demonstrating that these tokens contribute cumulatively to separation. This directional change confirms that the tokens identified by the contribution analysis are not merely numerically prominent but play a causal role in shaping the distance structure between authors (Table C6–C9)

\begin{samepage}
\noindent Table C6. Effect of removing top-contributing tokens on inter-author distance (Burrows’s Delta).
\begin{center}\begin{tabular}{lll}\hline
Removed Top-K & Distance (Before) & Distance (After)\\
10 & 0.429 & 0.420\\
50 & 0.429 & 0.394\\
100 & 0.429 & 0.367\\
\hline\end{tabular}\end{center}
\end{samepage}

\begin{samepage}
\noindent Table C7. Effect of removing top-contributing tokens on inter-author distance (Cosine Delta)
\begin{center}\begin{tabular}{lll}\hline
Removed Top-K & Distance (Before) & Distance (After)\\
10 & 0.816 & 0.794\\
50 & 0.816 & 0.736\\
100 & 0.816 & 0.687\\
\hline\end{tabular}\end{center}
\end{samepage}

\begin{samepage}
\noindent Table C8. Effect of removing top-contributing tokens on inter-author distance (Jensen–Shannon Divergence).
\begin{center}\begin{tabular}{lll}\hline
Removed Top-K & Distance (Before) & Distance (After)\\
10 & 0.057 & 0.055\\
50 & 0.057 & 0.049\\
100 & 0.057 & 0.045\\
\hline\end{tabular}\end{center}
\end{samepage}

\begin{samepage}
\noindent Table C9. Effect of removing top-contributing tokens on inter-author distance (Rank-Turbulence Divergence).
\begin{center}\begin{tabular}{lll}\hline
Removed Top-K & Distance (Before) & Distance (After)\\
10 & 0.223 & 0.220\\
50 & 0.223 & 0.214\\
100 & 0.223 & 0.208\\
\hline\end{tabular}\end{center}
\end{samepage}

\end{document}